\documentclass{article}

\usepackage{arxiv}

\usepackage[utf8]{inputenc} 
\usepackage[T1]{fontenc}    
\usepackage{hyperref}       
\usepackage{url}            
\usepackage{booktabs}       
\usepackage{amsfonts}       
\usepackage{nicefrac}       
\usepackage{microtype}      
\usepackage{lipsum}
\usepackage{graphicx}
\graphicspath{ {./Pic/} {./Figures/PDF/} {./Figures/PNG/} }

\usepackage[cmex10]{amsmath}
\usepackage{amssymb, bm, bbm}
\usepackage{algorithmic}
\usepackage{algorithm}
\usepackage{multirow}
\usepackage{balance}

\title{A Deep Equilibrium Network for Hyperspectral Unmixing\thanks{The work was supported by the National Natural Science Foundation of China under Grant 61701337.~\textit{(\#~Corresponding Author: Fei Zhu)}}}

\author{
  Chentong Wang \\
  Center for Applied Mathematics, KL-AAGDM\\
  Tianjin University\\
  Tianjin 300072, China \\
  \texttt{chentong\_wang@tju.edu.cn} \\
  \And
  Jincheng Gao \\
  Center for Applied Mathematics, KL-AAGDM\\
  Tianjin University\\
  Tianjin 300072, China \\
  \texttt{2023233003@tju.edu.cn} \\
  \And
  Fei Zhu$^\#$ \\
  Center for Applied Mathematics, KL-AAGDM\\
  Tianjin University\\
  Tianjin 300072, China \\
  \texttt{fei.zhu@tju.edu.cn} \\
  \And
  Jie Chen \\
  School of Artificial Intelligence\\
  Northwestern Polytechnical University\\
  Xi'an 710072, China \\
  National Engineering Laboratory for \\ Integrated Aero-Space-Ground-Ocean \\ Big Data Application Technology \\
  \texttt{dr.jie.chen@ieee.org} \\
}

\begin{document}
\maketitle

\begin{abstract}
Hyperspectral unmixing (HU) is crucial for analyzing hyperspectral imagery, yet achieving accurate unmixing remains challenging. While traditional methods struggle to effectively model complex spectral-spatial features, deep learning approaches often lack physical interpretability. Unrolling-based methods, despite offering network interpretability, inadequately exploit spectral-spatial information and incur high memory costs and numerical precision issues during backpropagation. To address these limitations, we propose DEQ-Unmix, which reformulates abundance estimation as a deep equilibrium model, enabling efficient constant-memory training via implicit differentiation. It replaces the gradient operator of the data reconstruction term with a trainable convolutional network to capture spectral-spatial information. By leveraging implicit differentiation, DEQ-Unmix enables efficient and constant-memory backpropagation. Experiments on synthetic and two real-world datasets demonstrate that DEQ-Unmix achieves superior unmixing performance while maintaining constant memory cost.
\end{abstract}

\keywords{Hyperspectral unmixing \and deep equilibrium models (DEQ) \and neural networks \and deep learning}

\section{Introduction}
\label{sec:intro}
Hyperspectral imagery provides rich spectral details but suffers from limited spatial resolution~\cite{R1}. Hyperspectral unmixing addresses this by extracting constituent pure material signatures (endmembers) and estimating their pixel-wise fractional proportions (abundances).
The tasks can be approached either sequentially or simultaneously. Early methods include geometric techniques for endmember extraction~\cite{VCA}, constrained regression for abundance estimation~\cite{FCLS}, and convex optimization with handcrafted priors or constraints~\cite{TV}. Nonnegative matrix factorization (NMF) is a well-established framework to simultaneously estimate both endmembers and abundances~\cite{NMF}.

Deep learning has achieved remarkable success in unmixing, with autoencoders (AEs) emerging as a widely adopted framework. Early AE methods performed pixel-wise encoding to infer abundances, while a decoder reconstructed spectra with network weights representing the estimated endmembers~\cite{AE}. Subsequent works integrated various network structures, such as convolutional neural networks (CNNs) \cite{CyCU, R21} and Transformer networks \cite{DeepTrans, UST, DTUNet}, to better exploit spectral-spatial information for accurate abundance estimation. However, despite efforts to incorporate underlying mixing mechanisms into network design \cite{R21, DTUNet, HapkeCNN}, AE-based methods are still criticized as ``black boxes'' lacking physical interpretability.

Recent efforts have focused on integrating optimization-inspired modeling with deep learning for greater interpretability.
This includes the plug-and-play (PnP) framework, which employs deep denoisers as implicit regularizers to improve unmixing~\cite{PNP,AERED}.   
Another approach is deep unrolling, which unfolds iterative solvers into feed-forward networks with learnable layers \cite{Unrolling}. For example, SNMF-Net~\cite{SNMF} unfolds sparse NMF algorithm, and \cite{UnrollPnP} incorporates deep denoisers into the unrolling scheme.
Nevertheless, these methods still have limitations.
The PnP-based unmixing \cite{PNP, AERED} relies on separately pre-trained denoisers, restricting their flexibility in exploiting data information. Unrolling unmixing networks often insufficiently model complex spectral-spatial dependencies~\cite{SNMF,Unroll-sparse}, and per-layer parameterization leads to high memory costs and numerical precision issues during backpropagation~\cite{DEQInverse}.

In this paper, we propose {DEQ-Unmix}, the first hyperspectral unmixing network to leverage the principles of deep equilibrium models (DEQ)~\cite{DEQ} to address the aforementioned challenges.
Starting from the linear mixing model (LMM) with sparsity regularization, we first derive a fixed-point iteration for the unsupervised unmixing problem, and then replace the gradient operator of the data reconstruction term with a trainable network designed to integrate the spectral-spatial information of the image via a convolutional architecture. Unlike unrolling-based methods, DEQ-Unmix embeds the forward process as a fixed-point iteration, enabling parameter sharing across an arbitrarily deep number of unfolded layers.  This avoids the high memory costs and numerical precision issues of a finite, per-layer parameterized architecture. 
By leveraging implicit differentiation, DEQ-Unmix enables efficient and constant-memory backpropagation. 
Experiments on synthetic and two real-world hyperspectral datasets demonstrate that DEQ-Unmix achieves competitive unmixing performance while maintaining constant memory cost.


\section{Methodology}
\label{sec:method}
\subsection{Problem Formulation}
DEQ-Unmix is based on the well-established linear mixing model (LMM)~\cite{R1}. To preserve the spectral-spatial structure for subsequent network modeling, we use its tensor form:
\begin{equation}
    \boldsymbol{\mathcal{Y}}=\boldsymbol{\mathcal{A}}\times_3 \mathbf{M}+\boldsymbol{\mathcal{N}},
    \label{LMM}
\end{equation}
where \( \boldsymbol{\mathcal{Y}} \in \mathbb{R}^{h \times w \times L} \) is the observed hyperspectral image, and the mode-3 product ($\times_3$) decomposes it into the endmember matrix $\mathbf{M} \in \mathbb{R}^{L \times R}$ and the abundance tensor $\boldsymbol{\mathcal{A}} \in \mathbb{R}^{h \times w \times R}$. The term \( \boldsymbol{\mathcal{N}} \) is the additive noise.
Based on the LMM~\eqref{LMM}, the conventional unmixing problem is formulated by minimizing the reconstruction error under a fixed norm 
(e.g., the Frobenius norm), while imposing regularization on the abundances and endmembers~\cite{R1, NMF}.

Instead of using a fixed metric, we design a data-driven optimization method where a learnable functional metric, $\mathcal{S}\left(\cdot\right)$, serves as the data reconstruction function to implicitly incorporate spectral-spatial priors in the hyperspectral image.
For a physically meaningful solution, we impose the abundance non-negative constraint (ANC), sum-to-one constraint (ASC), and endmember non-negativity, while enforcing $\ell_1$-based regularization on the abundances for sparsity. DEQ-Unmix performs unmixing in an unsupervised manner to jointly estimate both endmembers and abundances.
Thus, the optimization problem becomes
\begin{equation}\label{optimization}
    \begin{aligned}
        &\mathop{\arg\min}\limits_{\mathbf{M}\geq 0,\boldsymbol{\mathcal{A}}\geq 0}\mathcal{S}\left(\boldsymbol{\mathcal{A}}\times_3\mathbf{M}, \boldsymbol{\mathcal{Y}} \right)+\lambda\|\boldsymbol{\mathcal{A}}\|_1,\\
        & \text{s.t.} \sum_{r=1}^{R} \boldsymbol{\mathcal{A}}_{i,j,r} = 1, \quad \forall~i =1... h,\, j=1... w,
    \end{aligned}
\end{equation}
where $\lambda$ is a positive scalar that controls the sparsity level, $R$ is assumed to be known \textit{a priori}.

\subsection{A Fixed-point Solver for Abundance Tensor}
To solve the problem in \eqref{optimization}, we derive an iterative update for the abundance tensor $\boldsymbol{\mathcal{A}}$ via a fixed-point iteration. 
The endmember matrix $\mathbf{M}$ is set as a learnable weight matrix $\mathbf{W}$.
The update rule for $\boldsymbol{\mathcal{A}}$ is derived from a proximal gradient descent approach, which is given by
\begin{equation}
    \boldsymbol{\mathcal{A}}^{(k+1)}
    = \mathrm{Softmax}\left(
    \gamma\mathrm{ST}_{\eta\cdot\lambda}\left(\boldsymbol{\mathcal{A}}^{(k)}
    - \eta \, \nabla\mathcal{S}\left(
    \boldsymbol{\mathcal{A}}^{(k)} \times_3 \mathbf{W},\, \boldsymbol{\mathcal{Y}}\right)
    \times_3 \mathbf{W}^\top
    \right)
    \right),
    \label{eq3}
\end{equation}
where $\eta$ is the step size. 
We use the Softmax function to enforce both the  ANC and ASC, while $\gamma$ is a temperature-like parameter controlling the sharpness of the output distribution of Softmax~\cite{wang2021understanding}. The term ${\rm ST}_{\eta\cdot\lambda}\left(\cdot\right)$ is the {soft-thresholding operator}, which is defined element-wise as
${\rm ST}_{\eta\cdot\lambda}\left(x\right)={\rm sign}\left(x\right)\cdot{\max}(\big|x\big|-\eta\cdot\lambda,0)$. The sparsity scalar $\lambda$ is implemented as a trainable network parameter.

To fully leverage the spectral-spatial context in observed image, we replace the gradient operator $\nabla\mathcal{S}$ with a learnable network module, $\mathcal{G}_{\theta}\left(\cdot,\cdot\right):\boldsymbol{\mathcal{X}}\times \boldsymbol{\mathcal{X}}\rightarrow \boldsymbol{\mathcal{X}}$. Therefore, the update rule for the abundance tensor in~\eqref{eq3} becomes:
\begin{equation}\label{update}
    \boldsymbol{\mathcal{A}}^{(k+1)}
    = \mathrm{Softmax}\left(
    \gamma \mathrm{ST}_{\eta\cdot\lambda}\left(\boldsymbol{\mathcal{A}}^{(k)}
    - \eta \, \mathcal{G}_\theta\left(
    \boldsymbol{\mathcal{A}}^{(k)} \times_3 \mathbf{W}, \boldsymbol{\mathcal{Y}}\right)
    \times_3 \mathbf{W}^\top
    \right)
    \right).
\end{equation}
The update of $\boldsymbol{\mathcal{A}}^{(k+1)}$ depends only on the previous iterate $\boldsymbol{\mathcal{A}}^{(k)}$, thus formulating a fixed-point iteration. Therefore, \eqref{update} can be written in a compact form:
\begin{equation}\label{fixed-point}
    \boldsymbol{\mathcal{A}}^{(k+1)}=f_{\Theta}\left(\boldsymbol{\mathcal{A}}^{(k)}, \boldsymbol{\mathcal{Y}}\right),
\end{equation}
where $f_{\Theta}\left(\cdot\right)$ is the equilibrium layer designed according to the update rule in \eqref{update}, and $\Theta=\{\theta, \mathbf{W}, \lambda\}$ includes all trainable parameters.
Specifically, the iteration in \eqref{fixed-point} terminates when the index $k$ exceeds the maximum iteration $K_{\max}$ or the residual $\|\boldsymbol{\mathcal{A}}^{(k+1)}-\boldsymbol{\mathcal{A}}^{(k)}\|_2$ falls below the tolerance $\epsilon$.

\subsection{Network Architecture}
\begin{figure*}[h]
    \centering
    \includegraphics[trim=5mm 8mm 6mm 7mm, clip, width=6.2in]{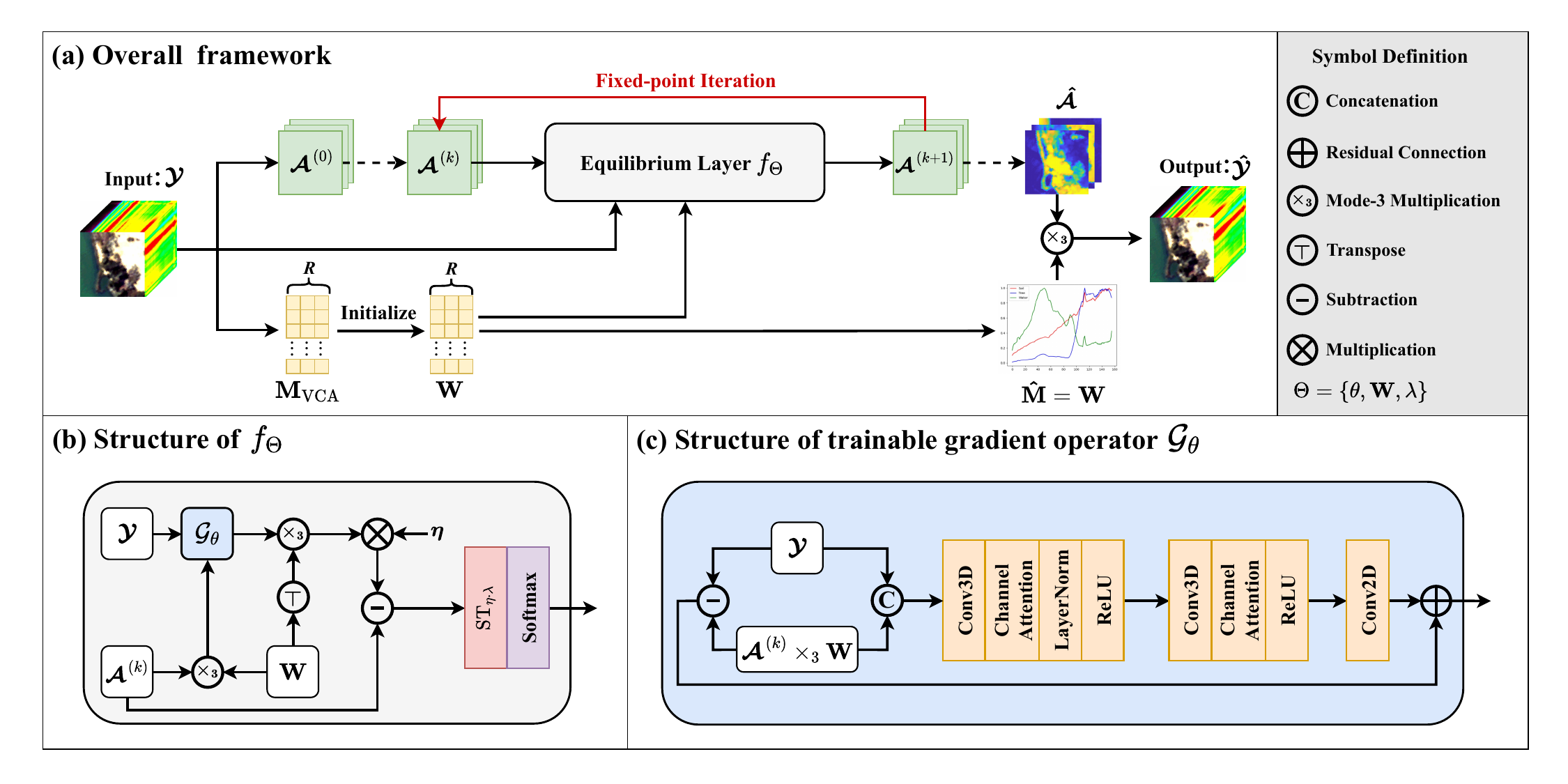}
    \caption{Overview of the proposed DEQ-Unmix Framework.}
    \label{DEQ-HU} 
\end{figure*}
The proposed DEQ-Unmix network's architecture is shown in Fig.~\ref{DEQ-HU}. For initialization, we use the endmember matrix $\mathbf{M}_{\text{VCA}}$ extracted by vertex component analysis (VCA)~\cite{VCA} to set $\mathbf{W}$, and obtain the initial abundance tensor $\boldsymbol{\mathcal{A}}^{(0)}$ using the corresponding solution from fully constrained least squares (FCLS)~\cite{FCLS}. The convolutional layers in $\mathcal{G}_{\theta}\left(\cdot, \cdot\right)$ adopt Xavier initialization, while $\lambda$ is set to $\lambda_0$.

As shown in~Fig.~\ref{DEQ-HU}(c), the network module $\mathcal{G}_{\theta}\left(\cdot, \cdot\right)$ is designed to approximate the gradient operator $\nabla\mathcal{S}$ while capturing spectral-spatial dependencies in the observed image by leveraging a specialized convolutional architecture. Its input is obtained by concatenating the observed image $\boldsymbol{\mathcal{Y}}$ and the current reconstruction $\boldsymbol{\mathcal{A}}^{(k)}\times_3\mathbf{W}$ along the channel dimension. The input is first processed by a block comprising a 3D convolutional layer with a $3\times3\times3$ kernel, a CBAM~\cite{CBAM}-inspired channel attention module, a Layer Normalization, and a ReLU activation function. This is followed by the second block, which consists of another 3D convolutional layer, a channel attention module, and a ReLU activation function. The feature map is then passed through a 2D convolutional layer with a $3\times3$ kernel, which reduces the channel dimension to align it with the observed image. Finally, a residual connection is applied between the difference $(\boldsymbol{\mathcal{A}}^{(k)}\times_3\mathbf{W}-\boldsymbol{\mathcal{Y}})$ and the output of the final layer, producing the output of $\mathcal{G}_{\theta}$.

According to \eqref{fixed-point}, the forward pass of the proposed DEQ-Unmix implements the following fixed-point iteration 
\begin{equation}\label{fixpoint-solution}
    \boldsymbol{\mathcal{A}}^{*}=f_{\Theta}\left(\boldsymbol{\mathcal{A}}^{*}, \boldsymbol{\mathcal{Y}}\right).
\end{equation}
Given the substantial computational challenges of solving this problem on high-dimensional hyperspectral data, we adopt Anderson acceleration \cite{anderson} to efficiently speed up the fixed-point iterations. The fixed-point solution serves as the estimated abundance tensor $\boldsymbol{\mathcal{\hat{A}}}$, while $\mathbf{W}$ functions as the estimated endmember matrix $\hat{\mathbf{M}}$. The image is reconstructed by
\begin{equation}
    \boldsymbol{\mathcal{\hat{Y}}}=\boldsymbol{\mathcal{\hat{A}}}\times_3 \hat{\mathbf{M}}.
\end{equation}
Following DEQ~\cite{DEQ}, we use the implicit function theorem to replace memory-intensive backpropagation with the efficient computation of an inverse Jacobian-vector product, given by
\begin{equation}\label{implict-backward}
    \frac{\partial\mathcal{L}}{\partial \Theta}={\frac{\partial f}{\partial \Theta}}^{\top}\left(\boldsymbol{\mathcal{I}}-\frac{\partial f}{\partial \boldsymbol{\mathcal{A}}^*}\right)^{-\top}\frac{\partial\mathcal{L}}{\partial \boldsymbol{\mathcal{A}}^*},
\end{equation}
where $\mathcal{L}$ is the loss function.
To approximate the inverse Jacobian-vector product, the tensor $\boldsymbol{\mathcal{V}}^*$ is defined as
\begin{equation}  \boldsymbol{\mathcal{V}}^*=\left(\boldsymbol{\mathcal{I}}-\frac{\partial f}{\partial \boldsymbol{\mathcal{A}}^*}\right)^{-\top}\frac{\partial\mathcal{L}}{\partial \boldsymbol{\mathcal{A}}^*}.
\end{equation}
Following~\cite{DEQInverse}, $\boldsymbol{\mathcal{V}}^*$ is the fixed point of the iteration given by
\begin{equation}\label{Jvb-iteration}
    \boldsymbol{\mathcal{V}}^{(k+1)}=\left(\frac{\partial f}{\partial \boldsymbol{\mathcal{A}}^*}\right)^{\top}\boldsymbol{\mathcal{V}}^{(k)}+\frac{\partial\mathcal{L}}{\partial \boldsymbol{\mathcal{A}}^*}.
\end{equation}
With the initial iterate $\boldsymbol{\mathcal{V}}^{(0)}=0$, the fixed-point $\boldsymbol{\mathcal{V}}^*$ is written as a Neumann series $\boldsymbol{\mathcal{V}}^*=\sum_{n=0}^{\infty}\left[\left(\frac{\partial f}{\partial \boldsymbol{\mathcal{A}}^*}\right)^{\top}\right]^n \!\! \frac{\partial\mathcal{L}}{\partial \boldsymbol{\mathcal{A}}^*}$, which is efficiently computed in practice using PyTorch.

\subsection{Loss Function}
The DEQ-Unmix framework is trained using two complementary losses: reconstruction error (RE) and spectral angle distance (SAD) loss. These are defined as follows:
\begin{equation}
    \mathcal{L}_{\text{RE}}(\boldsymbol{\mathcal{Y}}, \boldsymbol{\mathcal{\hat{Y}}}) = \frac{1}{N} \|\boldsymbol{\mathcal{\hat{Y}}}- \boldsymbol{\mathcal{Y}}\|_{F}^{2},
\end{equation}
\begin{equation}
    \mathcal{L}_{\text{SAD}}(\boldsymbol{\mathcal{Y}}, \boldsymbol{\mathcal{\hat{Y}}}) = \frac{1}{N} \sum_{i,j} \arccos \left( \frac{\langle \boldsymbol{\mathcal{Y}}_{i,j,:}, \boldsymbol{\mathcal{\hat{Y}}}_{i,j,:} \rangle}{\|\boldsymbol{\mathcal{Y}}_{i,j,:}\|_{2} \|\boldsymbol{\mathcal{\hat{Y}}}_{i,j,:}\|_{2}} \right),
\end{equation}
where $N={h \times w} $ is the number of pixels. 
The RE loss enforces pixel-wise fidelity but is sensitive to spectral scaling. To address this, the SAD loss measures angular similarity between spectra and is inherently scale-invariant under uniform scaling of endmembers. 
Following \cite{DeepTrans}, we define the total loss function as a weighted sum of the two losses
\begin{equation}\label{loss}
    \mathcal{L}=\alpha \mathcal{L}_{\text{RE}}+\mathcal{L}_{\text{SAD}},
\end{equation}
where $\alpha$ is a hyperparameter that balances the two losses.

\section{Experimental Results}
\label{sec:experiments}
In this section, we conduct experiments on both synthetic and real hyperspectral datasets. We compare DEQ-Unmix with several related unmixing techniques: VCA~\cite{VCA}+FCLS~\cite{FCLS}, PnP-BM3D~\cite{PNP}, PnP-BM4D~\cite{PNP}, AE-RED~\cite{AERED}, the Transformer-based DeepTrans~\cite{DeepTrans}, and the unrolling-based SNMF-Net~\cite{SNMF}. For a fair comparison, all methods use the same VCA initialization, and their parameters follow the procedures in the original papers.

The unmixing performance is evaluated by the abundance root mean square error (RMSE), ${\rm aRMSE}=({\frac{ \| \bm{\mathcal{\hat{A}}}-\bm{\mathcal{A}} \|_{F}^{2}}{NR}})^{1/2}$, and the endmember mean spectral angle distance (mSAD), ${\rm mSAD}=\frac{1}{R}\sum_{r=1}^{R}\arccos\left(\frac{\langle \hat{\mathbf{m}}_{r}, \mathbf{m}_{r} \rangle}{\|\hat{\mathbf{m}}_{r}\|_{2}\|\mathbf{m}_{r}\|_{2}}\right)$. Notably, each experiment is repeated ten times and average unmixing results are reported.

\subsection{Experiments with Synthetic Datasets}
We generated two synthetic datasets using the Hyperspectral Imagery Synthetic toolbox \cite{borsoi2021spectral}, each of size $100\times 100$ pixels. We selected $R=6$ endmembers with $L=224$ bands from the USGS spectral library. The corresponding abundance maps were generated via Gaussian random fields, with values capped at 0.85 to simulate highly mixed scenarios. Gaussian noise was added to the data to simulate realistic conditions, with signal-to-noise ratio (SNR) levels of 15 dB and 30 dB.

For synthetic datasets, we adopted consistent global settings: $\lambda_{0}=0.01$, $K_{\max}=10$, $\eta=0.04$, $\alpha=1$, and a weight decay of $10^{-5}$. The base learning rate was set to $0.01$ for all parameters except $\mathbf{W}$. Scene-specific adjustments were made based on noise levels: for SNR=15 dB, we set the learning rate of $\mathbf{W}$ to $0.003$ and $\gamma=0.9$; for SNR=30 dB, these were adjusted to $0.005$ and $0.8$, respectively.

Quantitative results in Table \ref{syn_result} show that DEQ-Unmix achieves competitive performance under both noise conditions. Notably, even in the challenging high-noise scenario (SNR=15 dB), our model maintains effective unmixing performance, demonstrating its robustness to noise.

\begin{table}[h]
    \caption{Unmixing Results on Synthetic Datasets. Best Results in Bold.}
    \renewcommand\arraystretch{0.9}
    \centering
    \small
    \begin{tabular}{c|cc|cc}
        \toprule[1pt]
        \multirow{2}{*}{Method} & \multicolumn{2}{c|}{SNR=15 dB}    & \multicolumn{2}{c}{SNR=30 dB}     \\ \cline{2-5} 
        & ${\rm aRMSE}$   & ${\rm mSAD}$    & ${\rm aRMSE}$   & ${\rm mSAD}$    \\ \midrule[1pt]
        VCA+FCLS                & 0.0733          & 0.0824          & 0.0609          & 0.0751          \\
        PnP-BM3D                & 0.0586          & 0.0621          & 0.0552          & 0.0598          \\
        PnP-BM4D                & 0.0561          & 0.0601          & 0.0536          & 0.0562          \\
        DeepTrans               & 0.0499          & 0.0569          & 0.0485          & 0.0541          \\
        AE-RED                  & 0.0565          & 0.0575          & 0.0503          & 0.0546          \\
        SNMF-Net                & 0.0612          & 0.0581          & 0.0499          & 0.0577          \\
        DEQ-Unmix               & \textbf{0.0470} & \textbf{0.0465} & \textbf{0.0441} & \textbf{0.0452} \\ \bottomrule[1pt]
    \end{tabular}
    \label{syn_result}
\end{table}

\subsection{Experiments with Real Datasets}
The widely used real datasets are Samson and Apex.
The Samson dataset, collected by the SAMSON sensor~\cite{samson}, is a $95 \times 95$ sub-image with 156 bands and contains three endmembers (Soil, Tree, and Water). The Apex dataset, collected by the APEX sensor~\cite{apex}, is a $110 \times 110$ sub-image with 285 bands, containing four endmembers (Water, Tree, Road, and Roof). The Apex dataset exhibits more complex spatial textures and spectral characteristics.
For both datasets, ground truth (GT) endmembers and their corresponding reference abundances provided by~\cite{DeepTrans} are used to assess performance.

For both datasets, we set the initial parameter $\lambda_{0}=0.1$, the maximum iteration $K_{\max}=10$, and the step size $\eta=0.01$. The remaining parameters were tuned for each dataset. On Samson, the learning rate was $0.01$ for all parameters except $\mathbf{W}$, which used $0.006$. A fixed weight decay of $10^{-5}$ was applied to all parameters, with $\gamma=1$ and $\alpha=0.1$. For Apex, the learning rate was $0.006$ for all parameters except for $\mathbf{W}$, which used $0.01$. The weight decays were $0$ for $\mathbf{W}$ and $10^{-3}$ for the other parameters, with $\gamma=1.2$ and $\alpha=10$.

\begin{figure}[h!]
    \centering
    \includegraphics[trim=0mm 0mm 0mm 0mm, clip, width=5in]{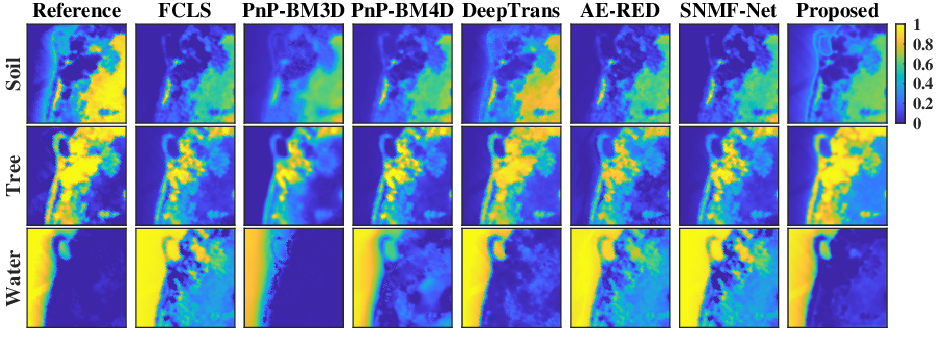}
    \caption{Estimated abundance maps for the Samson dataset.}
    \label{AbunSamson}
\end{figure}

\begin{figure}[h!]
    \centering
    \includegraphics[trim=0mm 0mm 0mm 0mm, clip, width=5in]{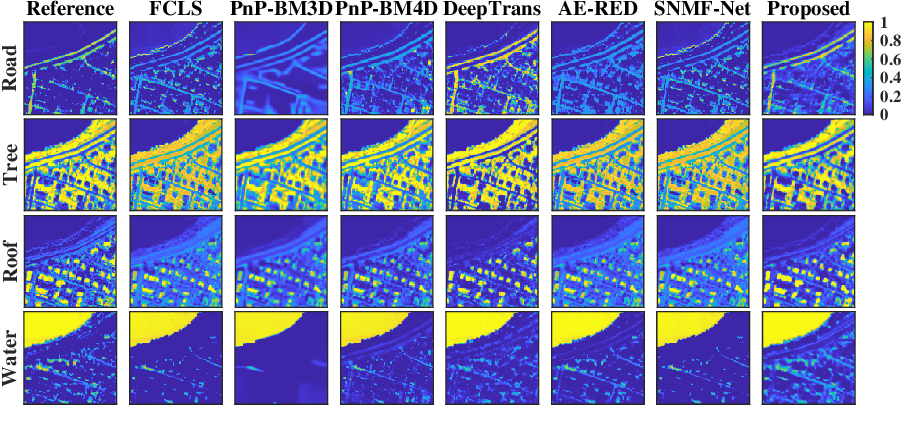}
    \caption{Estimated abundance maps for the Apex dataset.}
    \label{AbunApex}
\end{figure}

Quantitative results are reported in Table \ref{real_result}.
On the Samson dataset, the proposed DEQ-Unmix achieves the best performance in both abundance and endmember estimation. The Apex dataset presents greater challenges due to its richer spectral information and more complex spatial structures. Nevertheless, our method still achieves the best results in abundance estimation and a competitive performance in endmember estimation, ranking second to DeepTrans. Visual comparisons of the estimated abundance maps on Samson and Apex are shown in Fig.~\ref{AbunSamson} and Fig.~\ref{AbunApex}, respectively. These comparisons demonstrate that DEQ-Unmix estimates abundance maps that are visually aligned with the reference GT.

\begin{table}[h]
    \caption{Unmixing Results on Samson and Apex. Best Results in Bold.}
    \renewcommand\arraystretch{0.9}
    \centering
    \small
    \begin{tabular}{c|cc|cc}
        \toprule[1pt]
        \multirow{2}{*}{Method} & \multicolumn{2}{c|}{Samson} & \multicolumn{2}{c}{Apex} \\
        \cline{2-5}
        & $\rm{a}{\rm RMSE}$ & $\rm{m}{\rm SAD}$ & $\rm{a}{\rm RMSE}$ & $\rm{m}{\rm SAD}$ \\
        \midrule[1pt]
        VCA+FCLS & 0.2835 & 0.0667 & 0.1581 & 0.3958 \\
        PnP-BM3D & 0.1873 & 0.0362 & 0.1466 & 0.2023 \\
        PnP-BM4D & 0.1928 & 0.0656 & 0.1353 & 0.2463 \\
        DeepTrans & 0.0958 & 0.0448 & 0.1211 & \textbf{0.0993} \\
        AE-RED & 0.2633 & 0.0668 & 0.1214 & 0.2720 \\
        SNMF-Net & 0.2776 & 0.0641 & 0.1553 & 0.2528 \\
        DEQ-Unmix & \textbf{0.0621} & \textbf{0.0279} & \textbf{0.1077} & 0.1236 \\
        \bottomrule[1pt]
    \end{tabular}
    \label{real_result}
\end{table}

\subsection{Ablation Study}
We compare DEQ-Unmix against two unrolled baselines with identical depth ($K_{\text{max}}=10$) and the same network operator on the Samson dataset: Unroll (using distinct parameters per layer) and Unroll-S (sharing parameters across layers). As shown in Table \ref{ablation}, DEQ-Unmix significantly reduces memory consumption and training time compared to unrolled baselines, while achieving superior unmixing accuracy. This demonstrates that our equilibrium-based formulation enables efficient use of complex spectral-spatial modules without the memory overhead of standard unrolling.

\begin{table}[h!]
    \caption{Ablation results on the Samson dataset comparing Parameter Scale (MB), Running Time (s), Memory (GB), aRMSE, and mSAD. Best Results in Bold.}
    \setlength{\tabcolsep}{5.8pt}
    \renewcommand\arraystretch{0.9}
    \centering
    \footnotesize
    \begin{tabular}{c|ccccc}
        \toprule[1pt]
        Method   & \begin{tabular}[c]{@{}c@{}}Parameter\\ Scale\end{tabular} & \begin{tabular}[c]{@{}c@{}}Running\\ Time\end{tabular} & Memory           & ${\rm aRMSE}$   & ${\rm mSAD}$    \\ \midrule[1pt]
        Unroll   & 668.6323                                                  & 0.6155                                                 & 1.9317          & 0.1007          & 0.0531          \\
        Unroll-S & \textbf{66.8648}                                                  & 0.5695                                                 & 1.3428          & 0.0979          & 0.0506          \\
        DEQ-Unmix & \textbf{66.8648}                                           & \textbf{0.4615}                                        & \textbf{0.2595} & \textbf{0.0621} & \textbf{0.0279} \\ \bottomrule[1pt]
    \end{tabular}
    \label{ablation}
\end{table}

\section{Conclusion}
\label{sec:conclusion}
We propose DEQ-Unmix, a deep equilibrium hyperspectral unmixing framework that integrates physical modeling with deep learning. It reformulates the abundance estimation problem as a fixed-point iteration, with a trainable convolutional network replacing the traditional gradient operator of the data reconstruction term. This fixed-point formulation enables parameter sharing across layers, while the network design captures spectral-spatial dependencies. By leveraging implicit differentiation, DEQ-Unmix provides efficient and constant-memory backpropagation. Experiments on synthetic datasets, the Samson dataset, and the Apex dataset demonstrate the unmixing effectiveness of the proposed DEQ-Unmix.

\balance
\bibliographystyle{unsrt}
\bibliography{refs}

\end{document}